\documentclass{article}

\usepackage{color}
\usepackage{apacite}
\usepackage{graphicx}
\usepackage[english]{babel}

\begin{document}

\title{Reply to the commentary ``Be careful when assuming the obvious'', by P. Alday}

\author{Ramon Ferrer-i-Cancho}

\date{}

\maketitle 

\begin{abstract}
Here we respond to some comments by Alday concerning headedness in linguistic theory and the validity of the assumptions of a mathematical model for word order. For brevity, we focus only on two assumptions: the unit of measurement of dependency length and the monotonicity of the cost of a dependency as a function of its length. We also revise the implicit psychological bias in Alday's comments. Notwithstanding, Alday is indicating the path for linguistic research with his unusual concerns about parsimony from multiple dimensions. 
\end{abstract}

{\bf Kewords: word order, headedness, principles and parameters theory, clitics, units of measurement, language evolution }

\section{Headedness}

\label{headedness_section}

The principle of dependency length minimization makes conditional predictions, 
which do not imply that placing heads at the center is optimal in general \cite{Ferrer2013e}. In the context of just one head and at least one dependent, the principle predicts that the central placement of the head is optimal when there are at least two dependents but that placement is irrelevant if there is only one dependent.  In the context of an initial verb followed by two arguments (e.g., subject and object), the principle predicts that the heads of the verbal arguments are placed first with respect to their dependents,  e.g., articles or adjectives follow the head noun, whereas, for a final verb, the prediction is that the heads of the arguments are placed last with respect to their dependents, e.g. articles or adjectives precede the head noun \cite{Ferrer2013e}. 

The principle of dependency length minimization predicts consistent branching, to some extent, for 
verb-first or verb-last placements.\footnote{Our notion of branching and related concepts is based on dependency grammar \cite{Melcuk1988}, which makes less arbitrary assumptions about possible branchings than phrase structure approaches.}
If there is one head, the principle predicts a rather symmetric head placement but various heads can lead to an anti-symmetric placement \cite{Ferrer2013e,Ferrer2008e}. Importantly, the anti-symmetric placements do not need to be consistent. Thus, in an SOV language the optimal placement of an (inflected) verbal auxiliary is after the verbal head \cite{Ferrer2008e} but the optimal placement of adjectives is before their nominal heads.\footnote{\label{verbal_auxiliaries_footnote}We are assuming that verbal auxiliaries are dependents of verbs, but it could be argued that it should be the other way around, as in certain approaches to dependency grammar \cite{Melcuk1988}. There is no straightforward solution to this problem \cite[p. 106]{Heine1993a}. 
For our dependency length minimization arguments, the direction of the dependency is irrelevant. What matters is whether the heads of the subject and the object attach to the auxiliary verb (the auxiliary verb is then the hub) or to the main verb (the main verb is then the hub). A model where verbal auxiliaries are the hubs poses some problems. 
First, the terms ``main verb'' and ``auxiliary verb'' suggest that the main verb has to be the head.    
Second, research on grammaticalization suggests that auxiliaries are unlikely to be the heads at advanced stages of evolution \cite[p. 106]{Heine1993a}. Third, information theory predicts that the dependencies of high frequency words such as auxiliaries are likely to have weaker links than those of the main verb \cite{Ferrer2002f}, thus reducing the chances that auxiliaries win the competition for becoming the hubs. Fourth, an explanation for the relative placement of verbal auxiliaries given the 
placement of the main verb \cite{Ferrer2013e,Ferrer2008e} would be lost. A theory covering all the phenomena reached originally by the principle of online memory minimization would be heavier. }
   
The inconsistency of anti-symmetry is a challenge for principles and parameters theory, e.g. \citeA{Baker2001a}. 
Amongst other parameters, the theory typically defines one for headedness, specifying whether a head should follow or precede its dependents. That general parameter is clearly insufficient for the inconsistency above, but also because SVO languages put their verb at the center (head at the center) but then tend to put adjectives after the noun (head first). A possible solution is defining at least two headedness parameters, one for verbal heads and another for nominal heads, which is not very parsimonious (and still problematic: SOV follows left-branching for S and O but right branching for verbal auxiliaries).\footnote{For reasons explained in footnote \ref{verbal_auxiliaries_footnote}, we are not following the principles and parameters tradition strictly because we consider that the main verb (not the auxiliary verb) is the hub to which the subject and the object connect.} Instead, certain theoretical linguists may wish to involve interactions with other parameters in the discussion (see \citeA{Baker2001a} for candidates). But is that parsimonious enough? 
An alternative solution is considering the placement of the verb as a parameter. From that single parameter we have shown that it is possible to infer head-final placement within verb arguments in SOV, head-first placement within verb arguments in SVO, head-first placement within verb arguments of VSO/VOS \cite{Ferrer2013e} and also head first for the placement of inflected auxiliaries in SOV and head-final  for their placement in VSO \cite{Ferrer2008e}. However, the predictive power of the position of the verb does not imply that this is a fundamental parameter in a universal grammar sense: verb position might be determined by evolutionary time (e.g., verb-last being more likely in early stages of evolution) or sentence length (e.g., SVO being more likely in languages where speakers produce more elaborate sentences or simply longer sentences) as explained by \citeA{Ferrer2014a}.
 
   
\section{The monotonic dependency between cognitive cost and distance} 

\citeA{Ferrer2013e} assumes that the cognitive cost of a dependency is a strictly monotonic function of its length. \citeA{Alday2015a} considers that this assumption may not be valid. 
For simplicity, let us measure the length of a dependency in words. 
The online memory cost of a sentence of $n$ words can be defined as 
\begin{equation}
D = (n-1) \sum_{i=1}^{n-1} p(d) g(d),
\end{equation}
where $n - 1$  is the number of edges of a syntactic dependency tree of $n$ vertices, $p(d)$ is the proportion of dependencies of length $d$ and $g(d)$ is the cognitive cost of a dependency 
of length $d$.\footnote{$(n-1)p(d)$ is the number of dependencies of length $d$.} Assuming that $D$ is minimized when both the syntactic dependency tree and the values of $p(d)$ and $g(d)$ are constant, one can apply arguments analogous to those of \citeA{Ferrer2012d} for compression, which predict that $g(d)$ should tend to be a decreasing function of $d$.\footnote{In this prediction, $D$ is an analog of $E$, the mean energetic cost of a repertoire or vocabulary; $p(d)$ is the analog of $p_i$, the probability of the $i$-th most frequent unit of the repertoire (statistical analyzes suggest that $p(d)$ is on average a decreasing function of $d$ in real sentences \cite{Ferrer2004b}, as $p_i$ tends to shrink as $i$ grows); $g(d)$ is the analog of $e_i$, the energetic cost of the $i$-th most frequent unit of the repertoire \cite{Ferrer2012d}. The minimization of $E$ with constant distributions for $p_i$'s and $e_i$'s predicts that $e_i$ cannot decrease as $i$ increases. Thus the minimization of $D$ with constant distributions for the values of $p(d)$ and the values of $g(d)$ predicts that $g(d)$ cannot decrease as $d$ increases. }

\citeA{Alday2015a} reviews the issue of anti-locality effects \cite{Lewis2006a}: in certain circumstances, longer dependencies are processed faster. We believe that this could happen for various reasons. First, anti-locality might be a consequence of dependency length minimization itself (this is reminiscent of the idea that anti-locality effects derive from artifacts of the memory architecture
 \cite{Lewis2006a}). In Section \ref{headedness_section}, we reviewed that dependency length minimization and central head placement are not equivalent: in SOV languages, the placement of adjectives before the nominal heads of S and O, which is anti-local, leads to shorter dependencies than a central placement of those nominal heads within the S and O constituent. In other words, anti-locality within a constituent can increase the locality of the whole. This is a very important point for psycholinguistic research when the dependency length of a few dependencies or even just one dependency is considered to be relevant to argue for or against dependency minimization \cite{Konieczny2000a,Levy2012a}. Standard psycholinguistic research on word order is reductionistic.  

Second, conflicts between word order principles are at the core of Ferrer-i-Cancho's word order models \cite{Ferrer2014a}: 
\begin{itemize}
\item
Dependency length minimization is in conflict with maximization of predictability, e.g., postponing an element maximizes its predictability.  
\item 
A preference for SOV may simply mean that dependency length minimization is not the strongest principle (in short sequences, online memory is less important). 
\end{itemize}
Thus, anti-locality effects do not invalidate {\em a priori} the principle of dependency length minimization. Faster processing when dependents are farther apart does not need to be caused by their separation.\footnote{This is a general point in linguistic theory: the failure to meet the law of brevity does not invalidate the principle of compression \cite{Ferrer2012d}.} Other principles may be at play: postponing the head is optimal for maximizing its predictability \cite{Ferrer2014a}. 

Third, notice that \citeA{Ferrer2013e} assumes that the cost of a dependency of length $d$ is $g(d)$. In that setup, the identity of the dependents is irrelevant.  
However, we may consider a more general definition of the cost of a dependency between two units $u$ and $v$: $g(u, v, d)$. Intuitively, one would expect that $g(u, v, d)$ is lower when $u$ and $v$ are more strongly correlated, as this should make their dependency more robust against interference or decay. Lower values of $g(u,v,d)$ may facilitate the involvement of other word order principles.

\begin{figure}[h]
\begin{center}
\includegraphics[scale = 0.8]{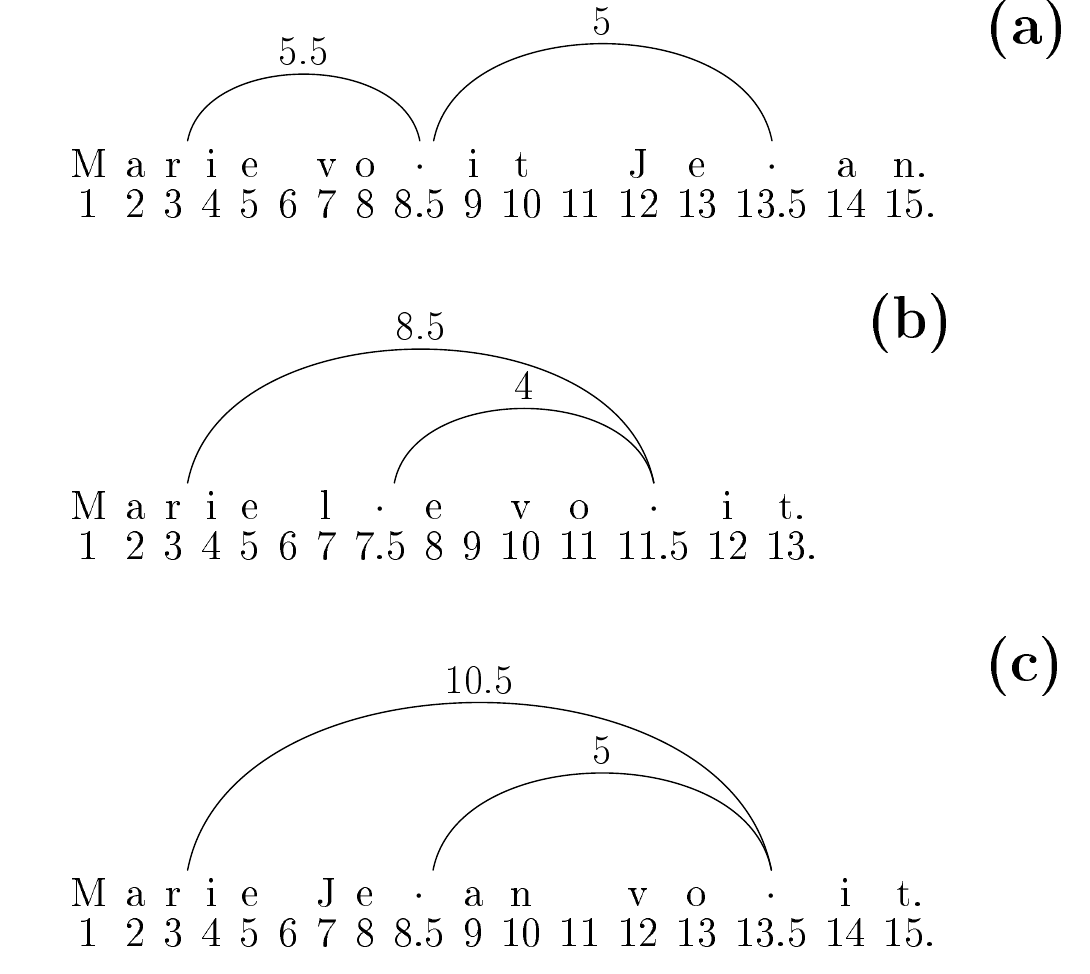}
\caption{\label{dependency_trees_figure} The syntactic dependency structure of three French sentences (the sentence in (c) is ``ungrammatical''). Edge labels indicate the length of the dependency in letters. Numbers below the words indicate the position of the word in the sentence from left to right. A dot is used to indicate the central position of words with an even number of letters.  }
\end{center}
\end{figure}

\section{The unit of measurement of dependency length}

A limitation of \citeA{Ferrer2013e} is that dependency length is measured in words. However, it could be measured in other linguistic units: syllables, morphemes, phonemes, etc. A higher level of precision might illuminate inconsistencies between the dominant order according to the criteria in \citeA{wals-s6} and other orders that appear recurrently in particular circumstances. \citeA{Alday2015a} points out the case of Italian, which is classified as SVO \cite{wals} but allows pronouns taking the role of the object to  appear before the verb. This is a common phenomenon occurring in other Romance languages (e.g., Catalan or French). Let us consider the case of French, which is 
 {\em ``SVO when the object is lexical, but
SOV when the object is prepositional''} \cite{Newmeyer2003a}, as in sentences (a) and (b) in Fig. \ref{dependency_trees_figure}.  
For simplicity, imagine that we measure dependency length in characters (this may make sense in a reading task, for instance). Imagine that the center of a word of length $\lambda$ is located at position $(\lambda+1)/2$ and that the space between two words counts as one character. 
For simplicity again, let us assume that dependencies originate from the center of the word and that the length of a dependency is the difference between the positions of the centers of the words involved. The sum of dependency lengths of the SOV sentence with a pronominal object (Fig. \ref{dependency_trees_figure} (b)) is in between that of the SVO sentence with nominal object (Fig. \ref{dependency_trees_figure} (a)) and that of SOV the sentence with nominal object (Fig. \ref{dependency_trees_figure} (c)). This is due to the brevity of the pronoun, suggesting that pressure for online memory minimization reduces if short words are involved. Another factor that may influence pressure for dependency length minimization is sentence 
length: dependency length minimization has been argued to be less necessary in short sentences, where the maximization of the predictability of the verb might be the winning principle \cite{Ferrer2014a}. Sentence length and word length might explain the tendency to adopt SOV in Romance languages when clitics are involved.  

\section{The psychological bias}

A wide majority of language researchers adopt a psychological perspective when dealing with word order and other linguistic phenomena \cite<e.g.,>{Konieczny2000a,Levy2012a, Alday2015a}. This perspective takes for granted that word order phenomena can ultimately be explained by principles (e.g., minimization of computational resources of the human brain) or mental processes that operate during sentence production or understanding within an individual, abstracting away from evolution (evolutionary history or evolutionary processes) and to a great extent from that individual's society. However, evidence that language structure or linguistic features are determined by social or political features is challenging this dominant view \cite{Lupyan2010a,Santacreu2013a}. 
\citeA{Ferrer2013e} also takes a psychological perspective when investigating the predictions of a principle of minimization of dependency length costs, but he departs from it by considering the possibility that certain word order configurations do not have a purely psychological origin but, rather, one that cannot be disentangled from evolution. For instance, the relative placement of the dependents within the subject or the object in SVO languages might be an adaptation that prevents regression to SOV \cite{Ferrer2013e}.
In terms of online memory minimization, an SVO language could place those dependents anywhere, but placements that are harder for SOV would be selected. Regarding dependency length minimization, SVO has more freedom to place dependents of the nominal heads after or before their heads than SOV does, but SVO has higher chances of survival in languages that place those dependents against the preferences of SOV.
Under this perspective, word order is regarded as a species that competes with other word orders.\footnote{Taking the argument further, one may argue that the competition is not at the surface level of word orders, but at the deep level of word order principles; in that case, word order principles would be like species that compete with other principles.} 
In sum, we may not be able to solve all word order puzzles through the organization of memory or mental processes involved in language production and understanding. Theorists may think that they have solved them with fatter theories requiring unnecessary or arbitrary parameters. Experimentalists may think that increasing the complexity of the experiments is the way to go. 

\section{Concluding remarks}

\citeA{Alday2015a} is very right about the need of parsimony in its multiple forms (simplicity, generality, theory-agnostic approach, unification of synchrony and diachrony, ...). Theoretical linguistics of the last century has chosen the easy path of modeling language  
by gratuitously adding parameters and then worrying {\em a posteriori} on constraints in number \cite{Hurford2012_Chapter3}\footnote{\citeA{Hurford2012_Chapter3} proposes wisely that those constraints derive from general cognitive constraints, such as dependency length minimization, to favor parsimony.} or evolution. Artificial intelligence researchers know very well that modeling is easy without any limit on the number of parameters or lacking any concern about the predictive capacity of the model on new data or new contexts (generality). 
The illusion of success caused by models that overfit linguistic phenomena and a blind belief in the autonomy of a discipline (against multidisciplinary research) prevent language specialists from incorporating simple and general models by 'outsiders'. 
We, as a community, should be wiser when we are not  satisfied by the limited fit of general models of language but do not care about overfitting.\footnote{'we' is used here to put myself in first place.} A lack of scientific depth can be hidden by abstract concepts and complex formalisms, but also by complicated psychological experiments or elaborate computer simulations. Modern model selection, which relies on a quantitative evaluation of models from a compromise between parsimony and goodness of fit \cite{Burnham2002a}, belongs to the future of language research. With his unusual concerns about parsimony, \citeA{Alday2015a} is reminding us that parsimony is more than an aesthetic requirement: rather, it is the key to a deep understanding of the myriad of issues with which he has cleverly challenged \citeA{Ferrer2013e}.

\section*{Acknowledgments}

We are very grateful to P. Alday for deep and detailed comments. The final version has benefited enormously from the comments and careful proofreading of S. Lambert. We also thank E. Santacreu-Vasut and K. Takashi-Ishi for helpful discussions. 
This work was supported by the grant BASMATI (TIN2011-27479-C04-03) from the Spanish Ministry of Science and Innovation.

\bibliographystyle{apacite}
\bibliography{../biblio/complex,../biblio/rferrericancho,../biblio/ling,../biblio/cs,../biblio/maths} 

\begin{thebibliography}{}

\bibitem[\protect\citeauthoryear{%
Alday%
}{%
Alday%
}{%
{\protect\APACyear{2015}}%
}]{%
Alday2015a}%
\APACinsertmetastar{%
Alday2015a}%
Alday, P.%
%
\unskip\
\newblock
\APACrefYearMonthDay{2015}{}{}.
\newblock
\BBOQ{}\APACrefatitle{Be careful when assuming the obvious. {Commentary} on
  "{The} placement of the head that minimizes online memory: a complex systems
  approach"}{Be careful when assuming the obvious. {Commentary} on "{The}
  placement of the head that minimizes online memory: a complex systems
  approach"}.\BBCQ{}
\newblock
\APACjournalVolNumPages{Language Dynamics and Change}{5}{1}{in press}.
\newblock
 \begin{APACrefURL} \url{http://arxiv.org/abs/1408.4753} \end{APACrefURL}
\PrintBackRefs{\CurrentBib}

\bibitem[\protect\citeauthoryear{%
Baker%
}{%
Baker%
}{%
{\protect\APACyear{2001}}%
}]{%
Baker2001a}%
\APACinsertmetastar{%
Baker2001a}%
Baker, M\BPBI C.%
%
\unskip\
\newblock
\APACrefYear{2001}.
\newblock
\APACrefbtitle{The atoms of language}{The atoms of language}.
\newblock
\APACaddressPublisher{New York}{Basic books}.
\PrintBackRefs{\CurrentBib}

\bibitem[\protect\citeauthoryear{%
Burnham%
\ \BBA{} Anderson%
}{%
Burnham%
\ \BBA{} Anderson%
}{%
{\protect\APACyear{2002}}%
}]{%
Burnham2002a}%
\APACinsertmetastar{%
Burnham2002a}%
Burnham, K\BPBI P.%
\BCBT{}\ \BBA{} Anderson, D\BPBI R.%
%
\unskip\
\newblock
\APACrefYear{2002}.
\newblock
\APACrefbtitle{Model selection and multimodel inference. {A} practical
  information-theoretic approach}{Model selection and multimodel inference. {A}
  practical information-theoretic approach}\ (\PrintOrdinal{2nd}\ \BEd).
\newblock
\APACaddressPublisher{New York}{Springer}.
\PrintBackRefs{\CurrentBib}

\bibitem[\protect\citeauthoryear{%
Dryer%
}{%
Dryer%
}{%
{\protect\APACyear{2013}}%
}]{%
wals-s6}%
\APACinsertmetastar{%
wals-s6}%
Dryer, M\BPBI S.%
%
\unskip\
\newblock
\APACrefYearMonthDay{2013}{}{}.
\newblock
\BBOQ{}\APACrefatitle{Determining Dominant Word Order}{Determining dominant
  word order}.\BBCQ{}
\newblock
\BIn{} M\BPBI S.~Dryer\ \BBA{} M.~Haspelmath\ (\BEDS), \APACrefbtitle{The World
  Atlas of Language Structures Online.}{The world atlas of language structures
  online.}
\newblock
\APACaddressPublisher{Leipzig}{Max Planck Institute for Evolutionary
  Anthropology}.
\newblock
 \begin{APACrefURL} \url{http://wals.info/chapter/s6} \end{APACrefURL}
\PrintBackRefs{\CurrentBib}

\bibitem[\protect\citeauthoryear{%
Dryer%
\ \BBA{} Haspelmath%
}{%
Dryer%
\ \BBA{} Haspelmath%
}{%
{\protect\APACyear{2013}}%
}]{%
wals}%
\APACinsertmetastar{%
wals}%
Dryer, M\BPBI S.%
\BCBT{}\ \BBA{} Haspelmath, M.%
\ (\BEDS).
\unskip\
\newblock
\APACrefYear{2013}.
\newblock
\APACrefbtitle{WALS Online}{Wals online}.
\newblock
\APACaddressPublisher{Leipzig}{Max Planck Institute for Evolutionary
  Anthropology}.
\newblock
 \begin{APACrefURL} \url{http://wals.info/} \end{APACrefURL}
\PrintBackRefs{\CurrentBib}

\bibitem[\protect\citeauthoryear{%
{Ferrer-i-Cancho}%
}{%
{Ferrer-i-Cancho}%
}{%
{\protect\APACyear{2004}}%
}]{%
Ferrer2004b}%
\APACinsertmetastar{%
Ferrer2004b}%
{Ferrer-i-Cancho}, R.%
%
\unskip\
\newblock
\APACrefYearMonthDay{2004}{}{}.
\newblock
\BBOQ{}\APACrefatitle{{Euclidean} distance between syntactically linked
  words}{{Euclidean} distance between syntactically linked words}.\BBCQ{}
\newblock
\APACjournalVolNumPages{Physical Review E}{70}{}{056135}.
\PrintBackRefs{\CurrentBib}

\bibitem[\protect\citeauthoryear{%
{Ferrer-i-Cancho}%
}{%
{Ferrer-i-Cancho}%
}{%
{\protect\APACyear{2008}}%
}]{%
Ferrer2008e}%
\APACinsertmetastar{%
Ferrer2008e}%
{Ferrer-i-Cancho}, R.%
%
\unskip\
\newblock
\APACrefYearMonthDay{2008}{}{}.
\newblock
\BBOQ{}\APACrefatitle{Some word order biases from limited brain resources. {A}
  mathematical approach}{Some word order biases from limited brain resources.
  {A} mathematical approach}.\BBCQ{}
\newblock
\APACjournalVolNumPages{Advances in Complex Systems}{11}{3}{393-414}.
\PrintBackRefs{\CurrentBib}

\bibitem[\protect\citeauthoryear{%
{Ferrer-i-Cancho}%
}{%
{Ferrer-i-Cancho}%
}{%
{\protect\APACyear{2014}}%
}]{%
Ferrer2014a}%
\APACinsertmetastar{%
Ferrer2014a}%
{Ferrer-i-Cancho}, R.%
%
\unskip\
\newblock
\APACrefYearMonthDay{2014}{}{}.
\newblock
\BBOQ{}\APACrefatitle{Why might {SOV} be initially preferred and then lost or
  recovered? {A} theoretical framework}{Why might {SOV} be initially preferred
  and then lost or recovered? {A} theoretical framework}.\BBCQ{}
\newblock
\BIn{} E\BPBI A.~Cartmill, S.~Roberts, H.~Lyn\BCBL{}\ \BBA{} H.~Cornish\
  (\BEDS), \APACrefbtitle{{THE EVOLUTION OF LANGUAGE - Proceedings of the 10th
  International Conference (EVOLANG10)}}{{THE EVOLUTION OF LANGUAGE -
  Proceedings of the 10th International Conference (EVOLANG10)}}\ (\BPG~66-73).
\newblock
\APACaddressPublisher{Vienna, Austria}{Wiley}.
\newblock
\APACrefnote{{Evolution} of {Language} {Conference} ({Evolang} 2014), April
  14-17}
\PrintBackRefs{\CurrentBib}

\bibitem[\protect\citeauthoryear{%
{Ferrer-i-Cancho}%
}{%
{Ferrer-i-Cancho}%
}{%
{\protect\APACyear{2015}}%
}]{%
Ferrer2013e}%
\APACinsertmetastar{%
Ferrer2013e}%
{Ferrer-i-Cancho}, R.%
%
\unskip\
\newblock
\APACrefYearMonthDay{2015}{}{}.
\newblock
\BBOQ{}\APACrefatitle{The placement of the head that minimizes online memory.
  {A} complex systems approach}{The placement of the head that minimizes online
  memory. {A} complex systems approach}.\BBCQ{}
\newblock
\APACjournalVolNumPages{Language Dynamics and Change}{5}{}{141-164}.
\newblock
\APACrefnote{http://arxiv.org/abs/1309.1939}
\PrintBackRefs{\CurrentBib}

\bibitem[\protect\citeauthoryear{%
{Ferrer-i-Cancho}%
\ \protect\BOthers{.}}{%
{Ferrer-i-Cancho}%
\ \protect\BOthers{.}}{%
{\protect\APACyear{2013}}%
}]{%
Ferrer2012d}%
\APACinsertmetastar{%
Ferrer2012d}%
{Ferrer-i-Cancho}, R.%
, Hern\'{a}ndez-Fern\'{a}ndez, A.%
, Lusseau, D.%
, Agoramoorthy, G.%
, Hsu, M\BPBI J.%
\BCBL{}\ \BBA{} Semple, S.%
%
\unskip\
\newblock
\APACrefYearMonthDay{2013}{}{}.
\newblock
\BBOQ{}\APACrefatitle{Compression as a universal principle of animal
  behavior}{Compression as a universal principle of animal behavior}.\BBCQ{}
\newblock
\APACjournalVolNumPages{Cognitive Science}{37}{8}{1565–1578}.
\PrintBackRefs{\CurrentBib}

\bibitem[\protect\citeauthoryear{%
{Ferrer-i-Cancho}%
\ \BBA{} Reina%
}{%
{Ferrer-i-Cancho}%
\ \BBA{} Reina%
}{%
{\protect\APACyear{2002}}%
}]{%
Ferrer2002f}%
\APACinsertmetastar{%
Ferrer2002f}%
{Ferrer-i-Cancho}, R.%
\BCBT{}\ \BBA{} Reina, F.%
%
\unskip\
\newblock
\APACrefYearMonthDay{2002}{}{}.
\newblock
\BBOQ{}\APACrefatitle{Quantifying the semantic contribution of
  particles}{Quantifying the semantic contribution of particles}.\BBCQ{}
\newblock
\APACjournalVolNumPages{Journal of Quantitative Linguistics}{9}{}{35-47}.
\PrintBackRefs{\CurrentBib}

\bibitem[\protect\citeauthoryear{%
Heine%
}{%
Heine%
}{%
{\protect\APACyear{1993}}%
}]{%
Heine1993a}%
\APACinsertmetastar{%
Heine1993a}%
Heine, B.%
%
\unskip\
\newblock
\APACrefYear{1993}.
\newblock
\APACrefbtitle{Auxiliaries. {Cognitive} forces and
  grammaticalization}{Auxiliaries. {Cognitive} forces and grammaticalization}.
\newblock
\APACaddressPublisher{New York}{Oxford University Press}.
\PrintBackRefs{\CurrentBib}

\bibitem[\protect\citeauthoryear{%
Hurford%
}{%
Hurford%
}{%
{\protect\APACyear{2012}}%
}]{%
Hurford2012_Chapter3}%
\APACinsertmetastar{%
Hurford2012_Chapter3}%
Hurford, J\BPBI R.%
%
\unskip\
\newblock
\APACrefYearMonthDay{2012}{}{}.
\newblock
\BBOQ{}\APACrefatitle{Chapter 3. {Syntax} in the light of evolution}{Chapter 3.
  {Syntax} in the light of evolution}.\BBCQ{}
\newblock
\BIn{} \APACrefbtitle{The origins of grammar. Language in the light of
  evolution II}{The origins of grammar. language in the light of evolution ii}\
  (\BPG~175-258).
\newblock
\APACaddressPublisher{Oxford}{Oxford University Press}.
\PrintBackRefs{\CurrentBib}

\bibitem[\protect\citeauthoryear{%
Konieczny%
}{%
Konieczny%
}{%
{\protect\APACyear{2000}}%
}]{%
Konieczny2000a}%
\APACinsertmetastar{%
Konieczny2000a}%
Konieczny, L.%
%
\unskip\
\newblock
\APACrefYearMonthDay{2000}{}{}.
\newblock
\BBOQ{}\APACrefatitle{Locality and parsing complexity}{Locality and parsing
  complexity}.\BBCQ{}
\newblock
\APACjournalVolNumPages{Journal of Psycholinguistic Research}{29}{}{627–645}.
\PrintBackRefs{\CurrentBib}

\bibitem[\protect\citeauthoryear{%
Levy%
, Fedorenko%
, Breen%
\BCBL{}\ \BBA{} Gibson%
}{%
Levy%
\ \protect\BOthers{.}}{%
{\protect\APACyear{2012}}%
}]{%
Levy2012a}%
\APACinsertmetastar{%
Levy2012a}%
Levy, R.%
, Fedorenko, E.%
, Breen, M.%
\BCBL{}\ \BBA{} Gibson, E.%
%
\unskip\
\newblock
\APACrefYearMonthDay{2012}{}{}.
\newblock
\BBOQ{}\APACrefatitle{The processing of extraposed structures in {English}}{The
  processing of extraposed structures in {English}}.\BBCQ{}
\newblock
\APACjournalVolNumPages{Cognition}{122}{1}{12 - 36}.
\PrintBackRefs{\CurrentBib}

\bibitem[\protect\citeauthoryear{%
Lewis%
, Vasishth%
\BCBL{}\ \BBA{} {Van Dyke}%
}{%
Lewis%
\ \protect\BOthers{.}}{%
{\protect\APACyear{2006}}%
}]{%
Lewis2006a}%
\APACinsertmetastar{%
Lewis2006a}%
Lewis, R.%
, Vasishth, S.%
\BCBL{}\ \BBA{} {Van Dyke}, J.%
%
\unskip\
\newblock
\APACrefYearMonthDay{2006}{}{}.
\newblock
\BBOQ{}\APACrefatitle{Computational principles of working memory in sentence
  comprehension}{Computational principles of working memory in sentence
  comprehension}.\BBCQ{}
\newblock
\APACjournalVolNumPages{Trends in Cognitive Sciences}{10}{}{447-454}.
\PrintBackRefs{\CurrentBib}

\bibitem[\protect\citeauthoryear{%
Lupyan%
\ \BBA{} Dale%
}{%
Lupyan%
\ \BBA{} Dale%
}{%
{\protect\APACyear{2010}}%
}]{%
Lupyan2010a}%
\APACinsertmetastar{%
Lupyan2010a}%
Lupyan, G.%
\BCBT{}\ \BBA{} Dale, R.%
%
\unskip\
\newblock
\APACrefYearMonthDay{2010}{}{}.
\newblock
\BBOQ{}\APACrefatitle{Language Structure Is Partly Determined by Social
  Structure}{Language structure is partly determined by social
  structure}.\BBCQ{}
\newblock
\APACjournalVolNumPages{PLoS ONE}{5}{1}{e8559}.
\PrintBackRefs{\CurrentBib}

\bibitem[\protect\citeauthoryear{%
Mel'\v{c}uk%
}{%
Mel'\v{c}uk%
}{%
{\protect\APACyear{1988}}%
}]{%
Melcuk1988}%
\APACinsertmetastar{%
Melcuk1988}%
Mel'\v{c}uk, I.%
%
\unskip\
\newblock
\APACrefYear{1988}.
\newblock
\APACrefbtitle{Dependency syntax: theory and practice}{Dependency syntax:
  theory and practice}.
\newblock
\APACaddressPublisher{Albany}{State of New York University Press}.
\PrintBackRefs{\CurrentBib}

\bibitem[\protect\citeauthoryear{%
Newmeyer%
}{%
Newmeyer%
}{%
{\protect\APACyear{2003}}%
}]{%
Newmeyer2003a}%
\APACinsertmetastar{%
Newmeyer2003a}%
Newmeyer, F\BPBI J.%
%
\unskip\
\newblock
\APACrefYearMonthDay{2003}{}{}.
\newblock
\BBOQ{}\APACrefatitle{Grammar is grammar and usage is usage}{Grammar is grammar
  and usage is usage}.\BBCQ{}
\newblock
\APACjournalVolNumPages{Language}{79}{}{682-707}.
\PrintBackRefs{\CurrentBib}

\bibitem[\protect\citeauthoryear{%
Santacreu-Vasut%
, Shoham%
\BCBL{}\ \BBA{} Gay%
}{%
Santacreu-Vasut%
\ \protect\BOthers{.}}{%
{\protect\APACyear{2013}}%
}]{%
Santacreu2013a}%
\APACinsertmetastar{%
Santacreu2013a}%
Santacreu-Vasut, E.%
, Shoham, A.%
\BCBL{}\ \BBA{} Gay, V.%
%
\unskip\
\newblock
\APACrefYearMonthDay{2013}{}{}.
\newblock
\BBOQ{}\APACrefatitle{Do female/male distinctions in language matter?
  {Evidence} from gender political quotas}{Do female/male distinctions in
  language matter? {Evidence} from gender political quotas}.\BBCQ{}
\newblock
\APACjournalVolNumPages{Applied Economics Letters}{20}{}{495-498}.
\PrintBackRefs{\CurrentBib}

\end{thebibliography}

\end{document}